\documentclass[letterpaper, 10 pt, conference]{ieeeconf}  

\IEEEoverridecommandlockouts                              

\usepackage{graphicx}
\usepackage{listings}
\title{\LARGE \bf
Towards Interactive, Incremental Programming of ROS Nodes*
\\\large Work in progress
}

\author{Sorin Adam$^{1}$ and Ulrik Pagh Schultz$^{2}$
\thanks{*This work was supported by the SAFE project}
\thanks{$^{1}$Sorin Adam is with Conpleks ApS, Struer, Denmark
        {\tt\small sorin.adam@conpleks.dk}}%
\thanks{$^{2}$Ulrik Pagh Schultz is with University of Southern Denmark, Odense, Denmark
        {\tt\small ups@mmmi.sdu.dk}}%
}

\begin{document}

\maketitle
\thispagestyle{empty}
\pagestyle{empty}

\begin{abstract}

Writing software for controlling robots is a complex task, usually demanding command of many programming languages and requiring significant experimentation. We believe that a bottom-up development process that complements traditional component- and MDSD-based approaches can facilitate experimentation.  We propose the use of an internal DSL providing both a tool to interactively create ROS nodes and a behaviour-replacement mechanism to interactively reshape existing ROS nodes by wrapping the external interfaces (the publish/subscribe topics), dynamically controlled using the Python command line interface.

\end{abstract}

\section{INTRODUCTION}

Writing software for controlling robots is a complex task, usually demanding a good command of one or more programming languages. As robotics is a multidisciplinary field, it is equally important to make accessible the software development for a larger number of roboticists with other expertise then software, as it is to increase the productivity of the software experts. Middleware such as ROS (Robot Operating System)~\cite{quigley:2009} or Orocos~\cite{Bruyninckx:01} are often used to simplify the development tasks, while model-driven software development (MDSD) or domain specific languages (DSLs) are used to increase the productivity~\cite{Sch11,Ste11,Ghe14}. 

Robotics software requires often significant experimentation, especially in the initial phase of development when debugging of algorithms and tuning of parameters prevail. Even if an overall software architecture described by a metamodel exists, adapting and customizing it for a given robot requires plasticity of the model and tools to reshape it. We believe that a bottom-up development process complementing MDSD while facilitating the initial iterations, is more appealing to the experimenter. As is the case with MDSD in general, we shorten the iteration time by using a DSL to generate the boilerplate code, thus providing ways for easy code modification. This enables the roboticist to focus on development by leaving the compliance with the restrictions of the programming environment to the tools. In essence, we propose a DSL supporting a bottom-up, experimental tinkering approach to development.

We propose the use of {\em component wrapping} model to combine the advantages of MDSD with an experimenter-friendly bottom-up approach of constructing the node model. Concretely, we propose to {\em create} ROS nodes using a more dynamic programming model and change the behaviour of existing ROS nodes by  {\em wrapping} the external interfaces (the publish/subscribe topics) while  reshaping them {\em interactively} using the Python command line interface. This way, existing functionality in a ROS node can be incrementally modified and experimented with, in effect allowing a kind of {\em wrappingnode} to be defined based on the functionality found in a {\em basenode}. We achieve that using a Python-based internal DSL language.

The rest of this paper is organised as follows: Section~\ref{sec:analysis} discusses practical problems associated with robotics software development and possible solutions, after which Section~\ref{sec:incrementalProgramming} presents our main contribution, last Section~\ref{sec:discussion} discusses the limitations of our approach as well as the possible extensions of its usage.

\section{PROBLEM ANALYSIS}
\label{sec:analysis}

We now present two concrete cases that motivate the work presented in this paper, followed by reflections on possible solutions.

\subsection{Case 1: Developing ROS nodes for FroboMind}

A ROS-based program is composed of nodes communicating through topics via a publish/subscribe mechanism\footnote{For the moment our work only addresses interaction through publish-subscribe, we plan to extend our approach to also include service calls, but this is left as future work.}. 
Experimenting requires often a small modification of an existing node. For example, if a new way of handling a message published on a topic is needed, several solutions are possible. If the source code is available, the options to alter the node functionality range  from in-place modification of the original code or cloning and changing the code, to usage of extension mechanisms of the node implementation language (e.g., a new C++ subclass if the original node structure permits). When the node source code is unavailable, the alternatives are often limited to re-write the complete node or interpose a new node on the message path and implementing the changes there. The last option is applicable even when the source code is available. From our experience with ROS-based FroboMind framework ~\cite{Jen12}, whenever the source code is available, the experimental tinkerer will often be tempted to use the clone-and-modify method. This approach introduces code redundancy and complicates the propagation of the original source code updates.

\subsection{Case 2: Experiments in safety restrictions}
\label{subsec:experiments-in-safety}

Enhancing the robot safety using software is a high interest research area for us. In a previous work, we have developed a safety-related DLS enforcing a set of safety rules to an existing robot~\cite{Adam:14}. Our experiments with the safety-related DSL required generation of robot misbehaviour in two different scenarios: when using a real robot (\textit{Frobit}~\cite{Larsen:2013}) and when experimenting with the simulated iClebo \textit{Kobuki} robot from the ROS distribution. While for the real robot, the fault simulation have been achieved by hardware means, for the simulated Kobuki a ROS node developed in Python has been created. Experimentation often triggered source code changes followed by restarts of the node. An environment where the node could be dynamically modified had the potential to accelerate the experiments. 

In general, writing software in Python is convenient for small scale development, as it is possible to experiment quickly using the command line interface. With the proper integration, the interactivity promoted by the Python development environment  could further enhance the runtime system interaction of the ROS command line tools. 

\subsection{Reflections on components wrapping and object-oriented inheritance}
\label{sec:background}

Our ROS experience indicates a need to incrementally and interactively modify or extend the behaviour of existing components by e.g., changing their response to messages, modifying messages, or communicating on new topics. Component-oriented programming suggests the use of wrappers~\cite{Gamma-al:95} or adapters~\cite{Hofmann:98} to address these needs. Such approaches are similar to a delegation-based model to inheritance~\cite{Ungar:87}, where an aggregate object receives method calls and can delegate them to a parent object.  Unlike inheritance, wrapping implies a significant runtime overhead. The components maintain separate identities, and moreover employ a black-box approach to composition where internal state is not visible, making fine-grained reuse difficult. Modular modification and extension of component behaviours through message interception has nevertheless been demonstrated as a useful programming model using composition filters~\cite{Bergmans:01}.  We observe that among these component adaptation techniques only inheritance is widely known and used by programmers, which suggests that presenting component adaptation using an wrapping model will facilitate practical use. Making the adaptation technique dynamic will enable interactive experimentation, well-known from object-oriented languages such as, e.g., Self and Smalltalk~\cite{Ungar:87,Goldberg:83}.

\section{INCREMENTAL AND INTERACTIVE ROS PROGRAMMING}
\label{sec:incrementalProgramming}
Our work proposes an incremental approach for altering ROS node functionality by using Python's command line interface as an interactive programming environment. This gives the experimenter a set of tools to work with a running system, especially useful during e.g., the initial phase of algorithm development and tuning, or hardware experimentation.

As a first step, we have developed an internal DSL implemented as a Python module. We use the DSL to enhance or modify already existing ROS nodes by using component wrapping mechanisms. In the end, the developed node encapsulates both the needed functionality and the node model, paving the way to higher level of MDSD usage in the robotics field by constructing the node model in a bottom-up, interactive approach, matching the way many roboticists develop robots.

\subsection{Concept}

\begin{figure}[t]
	\centering
		\includegraphics[width=0.5\textwidth]{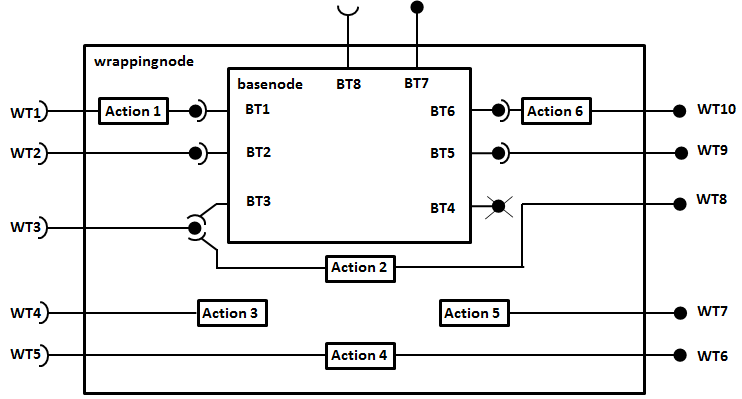}
	\caption{ROS node wrapping}\vspace*{-0.5cm} 
	\label{fig:overridingoverview}
\end{figure}

Our proposal for changing the functionality of an existing ROS node, applicable whether the source code is available or not, is to wrap the communication interfaces of an existing node, henceforth referred to as the {\em basenode }, via an interception layer referred to as the {\em wrappingnode}, allowing functionality to be modified or replaced. This approach is illustrated in Fig.~\ref{fig:overridingoverview} where the {\tt basenode } subscribing to the topics {\tt BT1, BT2, BT3, BT8} and publishing to the topics {\tt BT4, BT5, BT6, BT7}, has the communication interfaces consisting of the topics {\tt BT1, BT2, BT5, BT6}, wrapped through the {\tt wrappingnode}. The {\tt wrappingnode} not only wraps several communication interfaces of the {\tt basenode} via the  {\tt WT1, WT2, WT3, WT9, WT10} topics, but also adds new communication interfaces ( {\tt WT4, WT5, WT6, WT7} )and replaces one of the topics of the {\tt basenode} ({\tt BT4}) with {\tt WT8}. The blocks  {\tt Action1, Action2} and  {\tt Action6} implement the functionality changes for the {\tt basenode} topics {\tt BT1, BT4} and {\tt BT6}, while the blocks {\tt Action3, Action4} and  {\tt Action5} handle the new communication interfaces of the {\tt wrappingnode}. Note that wrapping unlike standard object-oriented inheritance, does not address the issue of the identity of the component (the {\tt wrappingnode} does not completely hide the {\tt basenode}), and moreover the state stored in the {\tt basenode} is not implicitly accessible to the {\tt wrappingnode}. In the end, by wrapping the {\tt basenode} we preserve parts of its functionality while also being able to extend or modify the original behaviour.

The wrapping approach impacts the MDSD development. If a robot architecture model exists and a new node must be added, node modelling is only required for the {\tt wrappingnode} as the {\tt basenode} is no longer visible as a separate entity in the new  model. On the other hand, if a {\tt basenode} model exists and the wrapping approach is applied, the model is altered in an invisible way for the {\tt basenode}. Therefore, the resulted {\tt wrappingnode} model could be seen as a reshaped version of the original one. 

\subsection{DSL by Example: Interactive ROS Node Creation}
\label{sec:exampleNodeCreation}

To provide an intuitive overview of the design of our DSL, we illustrate the succession of steps taken to create a new ROS node having the goal to drive the turtle from the ROS tutorials (Tutlesim) running in a circle. The code is presented as it is typed at Python's command line interface. As the presentation focus is on the DSL language, unimportant implementation details like importing the required libraries, are omitted.

First, the code implementing the callback functions implementing the behaviour of the node to the subscribed topics have to be typed:
\begin{lstlisting}[language=Python]
def showPose(data):
    print("Pose: {}".format(data))
\end{lstlisting}
Next, we declare the new node name:
\begin{lstlisting}[language=Python]
nd = rosNode("turtle_control_node")
\end{lstlisting}
It follows the node structure definition: the subscribe and published topics:
\begin{lstlisting}[language=Python]
{
nd.new.subscribe(topic = "/turtle1/pose", handler = showPose, msgType = Pose)
      .publish(topic = "/turtle1/cmd_vel", msgType = Twist)
}
\end{lstlisting}
We now {\em create} the defined ROS node. This operation causes the node to start running.
\begin{lstlisting}[language=Python]
nd.create()
\end{lstlisting}
As we wanted to make the Turtle in the Turtlesim to run in a circle, we will publish with 1Hz frequency the appropriate message to the {\em /turtle1/cmd\_vel} topic using a timer:
\begin{lstlisting}[language=Python]
def onTimer(event):
    msg = Twist()
    msg.linear.x = 2.0 
    msg.angular.z = 1.8
    nd.write("/turtle1/cmd_vel", msg )

timer = rospy.Timer(rospy.Duration(1), onTimer)
\end{lstlisting}

\subsection{DSL by Example: Kobuki Case Study}
The DSL could be used to interactively wrap an existing ROS node from the Python's command line interface. To exemplify this, we presnt the process of modifying one of the nodes used in the simulation demo example of section~\ref{subsec:experiments-in-safety}. As in the previous example, unimportant implementation details like importing the required libraries, are omitted. 

The interactive session starts when a {\em new node} {\\tt wnode} is  declared, {\em wrapping} an existing {\tt basenode} defined in a given package:
\begin{lstlisting}[language=Python] 
wnode = rosNode("experimental_move_base") 

{
wnode
   .baseNode("move_base") 
   .basePackage("move_base")            
}
\end{lstlisting}
Next, the {\tt wnode} structure of the topics being {\em reused} is declared as a pair of topic name and message type: 
\begin{lstlisting}[language=Python]  
{
 wnode.reuse
    .publish( topic = "cmd_vel", type = Twist)
    .publish(  topic =  "move_base/current_goal", type = PoseStamped)
    .publish( topic = "move_base/goal",  type = MoveBaseActionGoal)
    .subscribe(topic = "tf_static", type = TFMessage)
    .subscribe(topic = "move_base_simple/goal", type = PoseStamped)
    .subscribe(topic = "tf", type = TFMessage)
}
\end{lstlisting}
To {\em wrap} a {\tt basenode} topic, we relay the data between the {\tt subscribe} and {\tt publish} topics through a function and then we declare the {\tt new} topics: 
\begin{lstlisting}[language=Python]
def relayVelocity(data):
    wnode.write( "mobile_base/commands/velocity", data)
 
{
wnode.new
    .subscribe( topic = "cmd_vel", handler = relayVelocity, type = Twist)
    .publish( topic = "mobile_base/commands/velocity", type = Twist)
}
\end{lstlisting}
We now {\em create} the defined {\tt wnode}. This operation causes the node to start running.
\begin{lstlisting}[language=Python]  
wnode.create()
\end{lstlisting}
Now, the experimentation phase could begin. We first define a new function for controlling the speed of the robot through a global variable and then we {\em replace} the initial functionality:
\begin{lstlisting}[language=Python]
speed = 4.5

def controlVelocity(data):
    global speed
    if data.linear.x > 0:
        data.linear.x = speed 
    wnode.write( “mobile_base/commands/velocity”, data)

wnode.new
    .subscribe( topic = "cmd_vel", handler = controlVelocity, msgType = Twist)
\end{lstlisting}
From now on, the speed of the simulated robot could be controlled by simply changing the value of the global variable. The example triggers the maximum speed exceeded scenario by forcing the robot to run with 6m/s and exceeding in this way the defined maximum speed of 5m/s: 
\begin{lstlisting}[language=Python]
speed = 6
\end{lstlisting}

\subsection{Implementation}

The internal DSL is implemented as a Python library. In order to achieve the syntax form, the language implementation relies on method chaining. For improved readability, we prefer to split the chain into several lines, achieved by placing the code between either parentheses, square or curly brackets.  This code nesting  makes also possible unrestricted usage of indentation, further improving the code readability.  

The {\tt rosNode} class implements the {\tt basePackage}, {\tt baseNode}, {\tt write} and {\tt create} as well as it contains two instances of the {\tt topicHandler} class ({\tt reuse} and {\tt new}). The information related to the published and subscribed topics is encapsulated in dedicated classes ({\tt publisher} and {\tt subscriber}) instantiated trough the {\tt topicHandler} class and used by the {\tt subscribe}  and {\tt publish} methods.

The DSL is flexible. After declaring a new ROS node, the order of the {\tt wrappingnode} components is unimportant. Even more, after creation of the new node, it is possible to further modify it. The changes are dynamic, the node continues to run while topics are added, deleted or modified. Moreover, it is possible to replace on-the-fly any of the handling functions used by the node.

\subsection{Open issues}

The {\tt basenode} is launched by the DSL in a separate terminal command window when the {\tt create} method is called. As a result, two ROS nodes (the {\tt basenode} and the {\tt wrappingnode}) along with their internal topics, are visible to ROS commands like {\tt rosnode list} or {\tt rostopic list}.

When working with launch files, changes inside them are needed.  The modified node has to be removed from the original launch file and has to be created manually using the DSL via the Python's command line interface.

When the experimental work is concluded, the programmer normally wants to store the results and resume the work later. Also, when working in bigger projects involving several developers, it is desirable to share the source code among the team. None of these functionalities are  present in the current version of the DSL, but are left for future work. 

In our robotics research work, we use the FroboMind framework~\cite{Jen12}. As FroboMind does not utilize services, implementing them in the DSL was down-prioritized and left for future work. 

The direct support for timers is missing in the DSL, while the usage of configuration parameters is limited to the {\tt basenode}. However, if timers or configuration parameters for the {\tt wrappingnode} are needed, it is possible to implement them in the normal way used when developing ROS nodes in Python as exemplified in Section~\ref{sec:exampleNodeCreation}. 
 
\section{DISCUSSION}
\label{sec:discussion}

This section discusses the limitations of the DSL and proposes ways to address them, followed by possible enhancements of language and extensions to its current usage.

\subsection{Efficiency and its challenges}

While the DSL implementation allows creation of a new ROS node from scratch, without the need of any {\tt basenode}, the only practical side of it is to minimize the number of lines of code written, and group the ROS-related code in a way easier to understand. This could appeal to beginners, as learning ROS is neither fast nor easy, requiring both comprehension of the ROS concepts and mastering the ROS API. 

The DSL permits instrumenting an existing ROS node using a black box approach by first launching the node of interest, and then obtaining the information about the publish/subscribe topics together with the details about the messages used through standard ROS commands like {\tt rosnode info} or {\tt rostopic info}. After the {\tt wrappingnode} is created, the experimental phase could start.

As the DSL keeps an internal representation of the corresponding node model, it is possible to be save it in a format compatible with other MDSD tools like e.g., the external DSL developed for the ROS nodes used in FroboMind \cite{larsen:14}.

As the current DSL implementation internally relies on the usage of the standard ROS message transport, overhead is generated when the messages between the internal topics are serialized at sending time and deserialized at receiving time. This, of course, adds up delay in the transport of the message and increases the processor load. One solution to the problem would be to use the special ROS message {\tt anyMsg}, avoiding in this way the serialization-deserialization overhead. Alternatively, it could be possible to use nodelets. They are used in ROS to optimize the transport of messages by passing between them pointers to the buffers where the exchanged messages are stored. Optimizing the message transport is left for future work. 

\subsection{DSL design enhancements and usage extensions}

A first enhancement of the DSL as a tool is to extend it with a code generation part, able to produce Python or C++ code. Leaving the generation of the boilerplate type of code to the tool 
is appealing to both the inexperienced and the experienced ROS programmers. Moreover, the interactive way of experimenting with the ROS nodes makes possible to immediately experience the effect of any changes in the code as the DSL permits a live reshaping of a running node.

The DSL language syntax could be improved by changing the way the {\tt publish} and {\tt subscribe} topics are handled. For example, by using dynamic class attributes, publishing could be declared like:
\begin{lstlisting}
node.reuse.publish.cmd_vel(Twist) 
\end{lstlisting} 
rather than:
\begin{lstlisting} 
node.reuse.publish(topic="cmd_vel",type=Twist) 
\end{lstlisting}
The language syntax improvements are left as future work.

Another possible usage of the node wrapping concept implemented in our DSL is for safety enhancement purposes. Let's consider the case of an existing robot requiring to fulfil safety behaviour not originally implemented into it. While it is possible to enforce safety rules by adding dedicated ROS nodes (e.g. by using a safety-related DSL like in ~\cite{Adam:14}), the final reaction of the robot still depends of the original code, making the safety addition less effective. By developing the presented DSL to cover a safety-related scenario like this one, it will be possible to use the node wrapping for e.g., to safety wrap the actuator nodes and ensure the continuous control of the robot independent of the legacy software. Moreover, the node wrapping makes possible to use software of unknown provenance (SOUP) and ensure that, no matter how the original software reacts, the robot remains safe. Even more, the {\tt wrappingnode} could run on a different platform than the rest of the robot, where e.g. the response time of the node is guaranteed.

\addtolength{\textheight}{-12cm}   


\bibliographystyle{IEEEtran}
\bibliography{bibliography}

\end{document}